\documentclass[10pt,conference]{IEEEtran}
\IEEEoverridecommandlockouts
\usepackage[sort]{cite} 
\usepackage{amsmath,amssymb,amsfonts}
\usepackage{algorithmic}
\usepackage{graphicx}
\usepackage{textcomp}
\usepackage{xcolor}
\usepackage{comment}
\usepackage{flushend}
\DeclareMathOperator*{\argmax}{argmax}

\def\BibTeX{{\rm B\kern-.05em{\sc i\kern-.025em b}\kern-.08em
    T\kern-.1667em\lower.7ex\hbox{E}\kern-.125emX}}
\begin{document}
\title{Semantic Communication Enhanced by Knowledge Graph Representation Learning}
\title{Semantic Communication Enhanced by Knowledge Graph Representation Learning\\
\thanks{This work was supported by the 6G-GOALS project under the 6G SNS-JU Horizon program, n.101139232.}
}
\vspace{-16mm}
\author{\fontsize{1}{1}\selectfont
\IEEEauthorblockN{\fontsize{11}{12}\selectfont
Nour Hello\IEEEauthorrefmark{1}, Paolo Di Lorenzo\IEEEauthorrefmark{2}, Emilio Calvanese Strinati\IEEEauthorrefmark{1}
\vspace{0mm}}
\IEEEauthorrefmark{1}CEA-Leti, Université Grenoble Alpes, F-38000 Grenoble, France \\ \IEEEauthorrefmark{2} Sapienza University of Rome, Rome, Italy\\ \fontsize{8}{8}Emails: \{nour.hello, emilio.calvanese-strinati\}@cea.fr, paolo.dilorenzo@uniroma1.it
\vspace{-2mm}
}
\maketitle
\begin{abstract}
This paper investigates the advantages of representing and processing semantic knowledge extracted into graphs within the emerging paradigm of semantic communications. The proposed approach leverages semantic and pragmatic aspects, incorporating recent advances on large language models (LLMs) to achieve compact representations of knowledge to be processed and exchanged between intelligent agents. This is accomplished by using the cascade of LLMs and graph neural networks (GNNs) as semantic encoders, where information to be shared is selected to be meaningful at the receiver. The embedding vectors produced by the proposed semantic encoder represent information in the form of triplets: nodes (semantic concepts entities), edges(relations between concepts), nodes. Thus, semantic information is associated with the representation of relationships among elements in the space of semantic concept abstractions. In this paper, we investigate the potential of achieving high compression rates in communication by incorporating relations that link elements within graph embeddings. We propose sending semantic symbols solely equivalent to node embeddings through the wireless channel and inferring the complete knowledge graph at the receiver. Numerical simulations illustrate the effectiveness of leveraging knowledge graphs to semantically compress and transmit information.
\end{abstract}


\section{Introduction}

The recent advances in machine learning (ML) and Generative artificial intelligence (AI) are offering new fundamental enablers and integral components to design 6G connect-compute intelligent ecosystems \cite{Merluzzi2023} and new knowledge-centric applications based on the share of knowledge between (natural and/or artificial) intelligent agents. This has created the momentum for a new surge of interest in the paradigms of semantic and goal-oriented communications \cite{CalvaneseGOWSC2021,Kountouris21,gunduz2022beyond}.
Semantic communication transcends the paradigm of conventional communication by prioritizing the conveyance of semantic content over mere bit stream transmission. Indeed, while traditional communication systems are concerned with the technical tier of message exchange, semantic communication endeavors to operate within the semantic tier, encoding the meaning of information at the semantic encoder and interpreting an equivalent matching meaning at the semantic decoder \cite{weaver}. This approach to communication notably enhances the efficiency of distilling data for transmission and it improves the resilience against noise of the communication process.

There are three key pillars for semantic communications. First, as introduced by \cite{weaver}, the \textit{pragmatic} or \textit{goal-oriented} communications, in which the exchange of information is directed towards achieving specific objectives or goals. Semantic communications can be considered goal-oriented when the goal of the communications is not just to transmit data but to convey meaning or knowledge, where correct interpretation, manipulation, or composition of meaning aligns with predefined goals \cite{di2023goal}.  
To this end, several metrics have been proposed to evaluate the repercussions of semantic incongruences on the efficacy of decision-making processes, including the Age of Information (AoI) and the Value of Information (VoI) \cite{Kountouris21,b7}.
Second, in semantic communications, achieving effective and reliable knowledge sharing necessitates optimizing the compression of \textit{knowledge representation} to meet the reliability requirements of the application, as well as considering the context and goals of communication. In the literature, several solutions have been proposed to embed multi-modal data into a \textit{low-dimensional latent spaces},  relying on deep learning techniques like transformers \cite{sana}, generative adversarial networks \cite{gan}, graph convolutional neural network \cite {gnn}, LLMs \cite{llm_graphs}, just to name a few. The optimization of latent space compression should also ensure resistance against wireless and semantic noise, or mismatches between sources and destinations \cite{SanaSemEQ23}, as well as potential misalignment in the representation of (high-dimensional) latent spaces \cite{fiorellino2024dynamic}. This challenge differs from the extensively investigated mere semantic compression at semantic symbols level and semantic distillation \cite{Petar2021semantic}. We propose in this paper to approach it by using the cascade of LLMs and GNNs as semantic encoders. 
Third, the transmitter(s) and the receiver(s) are required to share and update a mutual knowledge base between them.
This calls for a new AI-native semantic architecture where the semantic plane, semantic RAN intelligent controllers (S-RIC), and semantic reasoning engine units are defined enabling new knowledge-centric applications \cite{Calvanese2024goal}.


In this paper, we focus on graph-based pragmatic semantic communications. Information extracted from data is semantically represented in knowledge graphs, formed by nodes (entities) and edges (relationships). Knowledge graphs provide structured representations of knowledge, organized into triples in the form of node-relationship-node, enabling the incorporation of semantic meaning into data. Thus, knowledge is structured and semantic messages selected, pragmatically aligning with the source's intent or the destination's objectives and context, or serving a specific goal \cite{Calvanese2024goal}. Recent works propose semantic communication systems based on the knowledge graph where semantic symbols are represented in the form of node-relationship-node triples for semantic extraction and interpretation \cite{b1,b3,b4,b5}. Those works mainly attempt to enhance accuracy, dispel interpretation ambiguities, improve data compression efficacy at the source, and enhance error correction robustness at the destination. Notably, \cite{b3} delineates an approach where transmission is adaptively modulated in response to channel quality, with a priority transmission accorded to semantically significant triples. Concurrently, \cite{b5} introduces a framework for cognitive semantic communication tailored to both single and multiple-user scenarios.

In the current state of the art, when semantic information is extracted in the form of Knowledge graphs, triples in the graphs are associated with semantic symbols to be transmitted \cite{b1,b3,b4,b5}. Thus, after (graph-based) semantic extraction, transmission signs (or patterns of signs) are optimized for transmission (semiotic \cite{morris2014writings}), but without exploiting the embedded meaning in the semantic message which is captured in the graph structure. With this paper, we propose to exploit the properties of the specific instance of the extracted graph such as its structure and descriptors of relations between nodes.
\\ \textbf{Contribution.} In this paper, we propose a novel end-to-end (E2E) pragmatic optimization semantic communications framework, optimizing the transmitter, the latent space representation and compression, and the receiver. This is done by representing semantic messages in the form of pragmatically sparsified knowledge graphs, being relevant to the receiver given the locally available knowledge. To learn the semantic relevant knowledge graph representation in lower-dimension latent spaces, we propose to cascade a pre-trained LLM module that associates embeddings to source data, with a GNN module that processes the combined output from the LLM module with the available inherent graphs topologies of the data. Such graphs are either available within the data set or can be extracted with LLM \cite{llm-kg}. The proposed semantic architecture encodes the knowledge graph in a batch of vectors, each one containing the semantic information about a node, the nodes connected to it, and the relations connecting them. The cardinality of such a batch of vectors encoding the graph is equal to the number of its nodes. Through this methodology, we numerically illustrate enhancements in both compression rates and communication robustness.

\section{System Model}
\begin{figure*}[htbp]
\centering
\includegraphics[width=0.75\textwidth, height=0.26\textheight]{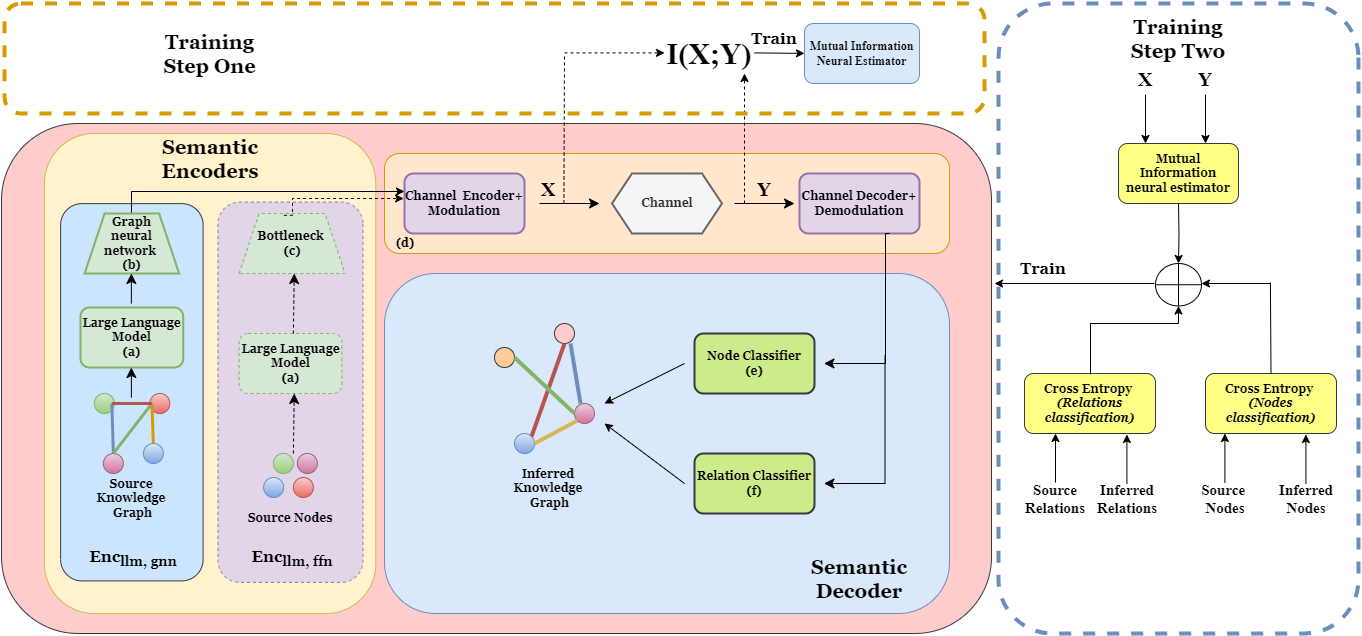}
\caption{System model of semantic representation learning on graphs}
\label{fig:diag}
\end{figure*}
The proposed architecture is described in Fig. ~\ref{fig:diag}. A knowledge graph is denoted as $\mathcal{G} = (\mathcal{N}, \mathcal{E})$, where $\mathcal{N} = \{\textit{n}_i\}_{i=1}^{\mathbf{N}_e}$ represents the set of nodes, where \(\textit{n}_i\) represents the textual attribute of the node with index $i$ and $\mathcal{E} = \{\textit{r}_{ij}\}_{i,j=1}^{\mathbf{N}_e}$ embodies the relational structure amongst these nodes. Here, $\textit{r}_{ij}$ signifies the relational attribute of the link directed from the source node of index $i$ to the target node of index $j$. The parameter $\mathbf{N}_e$ specifies the total number of nodes within the graph. The essential blocks of the system model are illustrated in the sequel.

\subsection{Semantic Encoders} \label{sec:semenc}
The semantic encoder associates the input knowledge graph $ \mathbf{g}$ with an array of low-dimensional embeddings $ \mathbf{x}^{\prime} = \{ \mathbf{x}^{\prime}_i\}_{i=1}^{ \mathbf{N}_e}$, such that each node of index $i$ of the source knowledge graph $ \mathbf{g}$ is associated with an embedding vector $ \mathbf{x}^{\prime}_i$.
We propose two semantic encoders named $\mathbf{Enc_{llm, gnn}}$ and $\mathbf{Enc_{llm, ffn}}$, respectively. In particular, $\mathbf{Enc_{llm, gnn}}$  cascades a pretrained LLM (Fig. ~\ref{fig:diag}, block (a))  with a GNN (Fig. ~\ref{fig:diag}, block (b)). The
LLM is adept at distilling textual features, consequently, we apply these models to synthesize the initial feature vectors pertinent to the textual attributes associated with nodes and relations of the knowledge graph. Those vectors constitute the preliminary node and relations feature vectors within GNN. Subsequently, the GNN is employed to synthesize node representations. This culminating representation is an amalgamation of textual and relational information pertinent to the node, encapsulating a comprehensive semantic representation. To be more specific, we have the following relations:
\begin{equation} \label{gnn_eq}
    \mathbf{x}_i = \mathbf{LLM}(\textit{n}_i), \hspace{0.2cm}
    \mathbf{e}_{j,i}= \mathbf{LLM}(\textit{r}_{ji}), \hspace{0.2cm}
\mathbf{x}^{\prime}_i =\mathbf{GNN}(\mathbf{X}_i, \mathbf{E}_{i}).
\end{equation}
Here, the LLM associates the initial features vectors $\mathbf{x}_i$, and $\mathbf{e}_{j,i}$ respectively with the descriptors of the node $\textit{n}_i$ and the relation $\textit{r}_{ji}$. Subsequently, the GNN generates a compact representation vector of the node of index $i$ by fusing $\mathbf{X}_i$ with $\mathbf{E}_{i}$, respective to the initial features of the neighboring nodes connected to the node $i$, and the initial features of the relations that culminate at the node $i$.
The intricate architecture of a knowledge graph, characterized by its attributed vertices and edges, necessitates the deployment of a graph neural network that must be adept at accommodating the inherent heterogeneity of the vertices, the directionality of the edges, and the rich feature representation of the edges. In this context, we exploit the graph isomorphism convolutional neural network in \cite{gine}, which reads as: 
 \begin{equation}
     \mathbf{x}^{\prime}_i = h_{\mathbf{\Theta}} \left( (1 + \epsilon) \cdot
\mathbf{x}_i + \sum_{j \in \mathcal{N}(i)} \mathrm{ReLU}
( \mathbf{x}_j + \mathbf{e}_{j,i} ) \right)
 \end{equation}
Here, $\mathbf{x}^{\prime}_i$ is the output features vector of node $i$, $h_{\mathbf{\Theta}}$ is a sequential neural network, and $\epsilon$ is a trainable value.
$\mathbf{x}_i$ is the node $i$'s initial features vector, and $\mathbf{e}_{j,i}$ are the features of the relational edge linking node $j$ to node $i$.

The second proposed encoder is denoted as $\mathbf{Enc_{llm, ffn}}$, and is composed by the cascade of a LLM with a feed-forward neural network (FFN) bottleneck (Fig. ~\ref{fig:diag}, block (c)). The aim of the FFN bottleneck step is to perform compression of the initial feature vector associated by the LLM to each node, but without taking into consideration the associative relationships in the knowledge graph. This aspect makes this encoder substantially different from $\mathbf{Enc_{llm, gnn}}$. Mathematically, the encoder produces the following vectors:
\begin{equation}
    \mathbf{x}_i = \mathbf{LLM}(\textit{n}_i), \qquad
    \mathbf{x}^{\prime}_i =\mathbf{FFN}(\mathbf{x}_i).
\end{equation}
Here, the LLM associates the initial features vector $\mathbf{x}_i$, to the node descriptor $\textit{n}_i$. Then, FFN compresses the node $i$'s representation via a (possibly tunable) bottleneck layer.

\subsection{Channel Coding}
The channel encoder consists of a FFN and a power normalization layer, modulating each low-dimensional vector $\mathbf{x}^{\prime}_i$ from the semantic encoder into $k$ complex symbols. The channel decoder, using an FFN architecture, demodulates the received symbols back into the set of latent space vectors  $\mathbf{y}= \{\mathbf{y}_i\}_{i=1}^{\mathbf{N}_e}$.
We use mutual information to determine the optimal channel encoding function (Fig. ~\ref{fig:diag}, block (d)) for an Additive White Gaussian Noise (AWGN) channel. The mutual information between the channel's input and output is approximated using the Mutual Information Neural Estimation (MINE) \cite{mine} framework.

\subsection{Semantic Decoder}
At the semantic decoder, the received message $\mathbf{\hat g}$  is decoded from a set of embedding vectors $\mathbf{y}= \{\mathbf{y}_i\}_{i=1}^{\mathbf{N}_e}$ to infer a knowledge graph congruent with the one intended by the source.
To infer the knowledge graph, we combine two decoding functions which separately classify the graph's nodes (Fig. ~\ref{fig:diag}, block (e)) and their relations (Fig. ~\ref{fig:diag}, block (f)).
 
\paragraph{Nodes Classifier} The node classification mechanism employs a deep architecture to categorize  $\mathbf{N}_e$ distinct decoded embedding vectors, attributing each to a specific node (i.e., semantic concept). In formulas, we have:
\begin{equation}\label{eq:mlp}
    \mathbf{z}_i = \text{MLP}(\mathbf{y}_i) 
    \end{equation}
    \begin{equation}
       \textit{n}_i =  \argmax_{i=1,\ldots,|\mathbb{E}|}\; \text{Softmax}(\mathbf{z}_i)
    \end{equation}
As delineated in \eqref{eq:mlp}, the architecture of the node decoder is modeled after a multi-layered perceptron (MLP) with skip connections, to associate a node with its attributed type over the set of nodes specified as $\mathbb{E}$.
\paragraph{Relation Classifier} The relation classifier works as a discerning mechanism to reconstruct the edges within the knowledge graph. It achieves this by utilizing node representations obtained from the graph encoder. Within this structure, combinations of embeddings are processed to associate commensurate relations. Specifically, we have the relations:
\begin{equation}
    (\mathbf{y}^{\prime}_i, \mathbf{y}^{\prime}_j) = \text{TransformerEncoder}(\mathbf{y}_i, \mathbf{y}_j),
\end{equation}
\begin{equation}
    \textit{r}_{ij} = \argmax_{k=1,\ldots,|\mathbb{R}|}\;  \text{Softmax} \left( \text{FC} (\mathbf{y}^{\prime}_i, \mathbf{y}^{\prime}_j) \right). 
\end{equation}
Overall, the relation decoder comprises a transformer encoder encapsulating self-attention heads, and feed-forward neural network transformation \cite{attention}. The transformer encoder takes the input vectors $\mathbf{y}_i$ and $\mathbf{y}_j$ associated with a combination of nodes of indices $i$ and $j$, and outputs new embedding vectors $\mathbf{y}^{\prime}_i$ and $\mathbf{y}^{\prime}_j$, which are subsequently input into a classification head (FC), that determines the relation between the examined combination through the corpus of relations designated by $\mathbb{R}$. This corpus includes the "none" relation indicating the absence of a direct relation between two nodes.

\subsection{Knowledge Graph Representation Learning}
Consider $\mathbf{\theta }$ and $\mathbf{\vartheta}$ as the respective trainable parameters of the semantic encoder and the semantic decoder. Drawing upon the insights of \cite{sana}, the loss function used to train the system under consideration writes as:
\begin{equation}
   \mathcal{L}_{{\mathbf{\theta }}.\vartheta }^{{\text{CE}}} - \alpha {I_{\mathbf{\theta }}}(X;Y)
\end{equation}
The function encapsulates the cross-entropy term augmented with a penalty parameter considering a weighted mutual information component, which reflects the interdependence between variables $X$ and $Y$ across the semantic channel from transmission to reception. The random variable $g$, intrinsic to a knowledge graph, manifests as the joint random variable outlining the stochastic properties of two independent random variables $n$ and $r$: the first-mentioned representing the nodes, and the second-mentioned standing for the relations. Overall, we have two parallel components: one dedicated to the inference of nodes, and the other designated on establishing relations between them. Mathematically, this can be cast as the minimization of the following function:
\begin{equation}
    \mathcal{L}_{\mathbf{\theta}, \vartheta}^{\text{CE}_{g}} = \mathcal{L}_{\mathbf{\theta}, \vartheta}^{\text{CE}_{n}} + \mathcal{L}_{\mathbf{\theta}, \vartheta}^{\text{CE}_{r}}
\end{equation}
One training epoch in our framework consists of two steps: Step one involves the calibration of the mutual information estimation model. Then, step two encompasses training the entire end-to-end (E2E) model using a dual-objective optimization strategy that combines cross-entropy loss minimization with mutual information maximization to improve predictive accuracy.
\begin{figure}[t]
\centerline{\includegraphics[height=0.25\textheight]{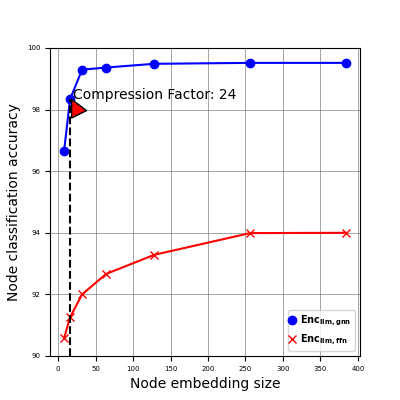}}
\caption{Node Classification Accuracy vs. Node Embedding Dimensionality.}
\label{res1}
\end{figure}

\section{Numerical Results}
We assess the performance of the proposed graph-based E2E wireless semantic communication system via numerical simulation. On the transmitter side, knowledge graphs of $N_e$ nodes are encoded into $N_e$ vectors of semantic embeddings, then each embedding vector is modulated into $k$ complex symbols to be transmitted over the AWGN channel. We compare the two encoders design which are detailed  in section \ref{sec:semenc}: $\mathbf{Enc_{llm, gnn}}$ and $\mathbf{Enc_{llm, ffn}}$. 
At the receiver, knowledge graphs are inferred from the received (noisy) semantic symbols. We assess the decoder's fidelity with the F1 score metric. The F1 score measures the average equivalence of nodes and relations between the transmitted and the decoded knowledge graphs by checking the exact match of the inferred triple elements with the source triple elements \footnote{https://github.com/WebNLG/WebNLG-Text-to-triples}. We compute the F1 score metric, utilizing the knowledge graphs from the WebNLG Project \cite{webnlg}. Training of the E2E system is performed at a reference signal-to-noise ratio (SNR) of 14 dB and a batch size of 8 graphs separately for both encoder designs. We use the pre-trained "all-MiniLM-L12-v2" LLM \cite{bert-sentence}, which produces feature vectors of size 384.

In Fig.~\ref{res1} we compare the performance in terms of node classification accuracy versus the embedding compression at the output of the encoders under noiseless wireless channel conditions. Our numerical results show that for an embedding compression factor of 24 or smaller $\mathbf{Enc_{llm, gnn}}$ exceed $98\%$ of node classification accuracy and outperforms $\mathbf{Enc_{llm, ffn}}$. Moreover, we observe that its performance just slightly depends on the embedding size used by the semantic communication system. This is thanks to the inclusion of graph topology information in $\mathbf{Enc_{llm, gnn}}$ and the superior efficacy of GNNs in condensing the semantics of graph nodes, compared with a feed-forward neural network.

\begin{figure}[t]
\centerline{\includegraphics[height=0.25\textheight]{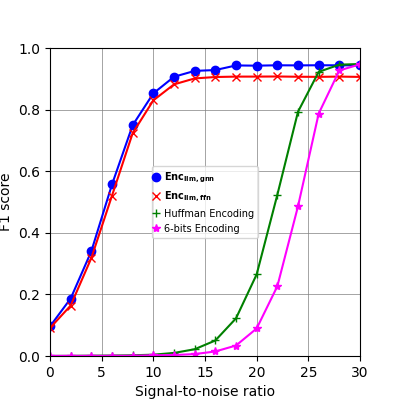}}
\caption{F1 Score of received and  source knowledge graphs matching vs. Signal To Noise Ratio when Dealing with AWGN Channel.}
\label{res2}
\end{figure}
In Fig.~\ref{res2}, we illustrate the behavior of the F1 score metric versus the SNR, considering an AWGN channel, and comparing the proposed encoders $\mathbf{Enc_{llm,gnn}}$ and $\mathbf{Enc_{llm, ffn}}$ with traditional encoding techniques: Huffman and 6-bits coding coupled with a 64-QAM modulation scheme. The output embedding size of both the GNN and the FFN bottleneck is set to 128. Each embedding vector is modulated into five complex symbols. As we can notice in In Fig.~\ref{res2}, both $\mathbf{Enc_{llm, gnn}}$ and $\mathbf{Enc_{llm, ffn}}$ performs closely the same in terms of F1 score, having $\mathbf{Enc_{llm, gnn}}$ outperforming of about 4\% compared to $\mathbf{Enc_{llm, ffn}}$ at high SNR regime. We observe how a graph-based semantic encoding and decoding notably outperforms the traditional techniques in the low SNR regime, requiring about 14 dBs less of SNR to reach their maximum F1 score. 
\begin{figure}[t]
\centerline{\includegraphics[height=0.25\textheight]{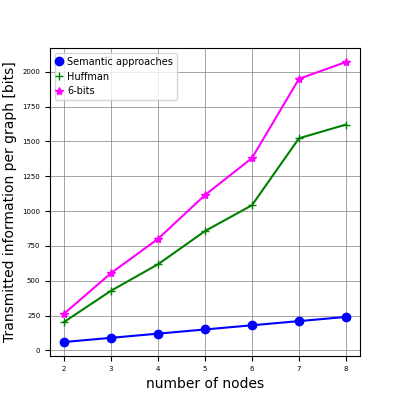}}
\caption{Transmitted information in bits per graph versus the number of nodes in the graph}
\label{compression}
\end{figure}

In Fig.~\ref{compression}, we present the transmitted information over the channel, measured as the average number of bits per knowledge graph, plotted against the number of nodes in the graph for the semantic approaches ($\mathbf{Enc_{llm, gnn}}$ or $\mathbf{Enc_{llm, ffn}}$), Huffman encoding, and 6-bits encoding. The average compression gain factor over the WebNLG dataset for the semantic approaches, with an embedding vector size of $128$ and modulating each vector to $5$ symbols, is $5.54$ compared to Huffman encoding and $7.17$ compared to 6-bit encoding.

\section{Conclusion}

In this paper, we investigate the effectiveness of LLM and GNN based encoders for semantic compression of knowledge graphs. To this end, we propose an end-to-end graph-based semantic communications framework. The goal is to embed knowledge graphs derived from data into a low-dimensional latent space, thus optimizing the compression of knowledge representation. We introduce a novel method for semantic encoding and decoding of knowledge graphs with two variants for semantic encoding. $\mathbf{Enc_{llm, gnn}}$ exploits the relational spectrum of nodes while $\mathbf{Enc_{llm, ffn}}$ treats each node isolated without considering its neighborhood. 


Our numerical results validate the effectiveness of our proposed framework. $\mathbf{Enc_{llm, gnn}}$ and $\mathbf{Enc_{llm, ffn}}$ reach respectively almost $99,5\%$ and $94\%$ of node classification accuracy. Moreover, $\mathbf{Enc_{llm, gnn}}$ enables a compression up to a factor of 24 for the nodes embedding size, while having node classification accuracy almost independent from the compressed embedding size. This is thanks to the inclusion of graph topology information in $\mathbf{Enc_{llm, gnn}}$ and the superior efficacy of GNNs in condensing the semantics of graph nodes. In addition, we also observe how the proposed graph-based semantic encoding and decoding functions notably outperform traditional Huffman and 6-bits coding in the low-average SNR regime, requiring about 14 dBs less of SNR to reach their maximum F1 score in our simulation settings. In conclusion, through the proposed design, we realize enhancements in both compression rates and communication robustness.


\bibliographystyle{IEEEtran} 
\bibliography{references} 

\begin{thebibliography}{10}
\providecommand{\url}[1]{#1}
\csname url@samestyle\endcsname
\providecommand{\newblock}{\relax}
\providecommand{\bibinfo}[2]{#2}
\providecommand{\BIBentrySTDinterwordspacing}{\spaceskip=0pt\relax}
\providecommand{\BIBentryALTinterwordstretchfactor}{4}
\providecommand{\BIBentryALTinterwordspacing}{\spaceskip=\fontdimen2\font plus
\BIBentryALTinterwordstretchfactor\fontdimen3\font minus \fontdimen4\font\relax}
\providecommand{\BIBforeignlanguage}[2]{{%
\expandafter\ifx\csname l@#1\endcsname\relax
\typeout{** WARNING: IEEEtran.bst: No hyphenation pattern has been}%
\typeout{** loaded for the language `#1'. Using the pattern for}%
\typeout{** the default language instead.}%
\else
\language=\csname l@#1\endcsname
\fi
#2}}
\providecommand{\BIBdecl}{\relax}
\BIBdecl

\bibitem{Merluzzi2023}
M.~Merluzzi \emph{et~al.}, ``The hexa-x project vision on artificial intelligence and machine learning-driven communication and computation co-design for 6g,'' \emph{IEEE Access}, vol.~11, pp. 65\,620--65\,648, 2023.

\bibitem{CalvaneseGOWSC2021}
E.~Calvanese~Strinati and S.~Barbarossa, ``{6G} networks: Beyond {S}hannon towards semantic and goal-oriented communications,'' \emph{J. Commun. Netw.}, vol. 190, February 2021.

\bibitem{Kountouris21}
M.~Kountouris and N.~Pappas, ``Semantics-empowered communication for networked intelligent systems,'' \emph{IEEE Communications Magazine}, vol.~59, no.~6, pp. 96--102, 2021.

\bibitem{gunduz2022beyond}
D.~G{\"u}nd{\"u}z \emph{et~al.}, ``Beyond transmitting bits: Context, semantics, and task-oriented communications,'' \emph{IEEE Journal on Selected Areas in Communications}, vol.~41, no.~1, pp. 5--41, 2022.

\bibitem{weaver}
\BIBentryALTinterwordspacing
W.~Weaver, ``Recent contributions to the mathematical theory of communication,'' \emph{ETC: A Review of General Semantics}, vol.~10, no.~4, pp. 261--281, 1953. [Online]. Available: \url{http://www.jstor.org/stable/42581364}
\BIBentrySTDinterwordspacing

\bibitem{di2023goal}
P.~Di~Lorenzo \emph{et~al.}, ``Goal-oriented communications for the iot: System design and adaptive resource optimization,'' \emph{IEEE Internet of Things Magazine}, vol.~6, no.~4, pp. 26--32, 2023.

\bibitem{b7}
\BIBentryALTinterwordspacing
A.~Li, S.~Wu, S.~Meng, R.~Lu, S.~Sun, and Q.~Zhang, ``Towards goal-oriented semantic communications: New metrics, open challenges, and future research directions,'' 2023. [Online]. Available: \url{https://arxiv.org/abs/2304.00848}
\BIBentrySTDinterwordspacing

\bibitem{sana}
M.~Sana and E.~C. Strinati, ``Learning semantics: An opportunity for effective 6g communications,'' in \emph{2022 IEEE 19th Annual Consumer Communications \& Networking Conference (CCNC)}, 2022, pp. 631--636.

\bibitem{gan}
M.~U. Lokumarambage, V.~S.~S. Gowrisetty, H.~Rezaei, T.~Sivalingam, N.~Rajatheva, and A.~Fernando, ``Wireless end-to-end image transmission system using semantic communications,'' \emph{IEEE Access}, vol.~11, pp. 37\,149--37\,163, 2023.

\bibitem{gnn}
K.~Tonchev, I.~Bozhilov, and A.~Manolova, ``Semantic communication system for 3d video,'' in \emph{2023 Joint International Conference on Digital Arts, Media and Technology with ECTI Northern Section Conference on Electrical, Electronics, Computer and Telecommunications Engineering (ECTI DAMT \& NCON)}, 2023, pp. 542--547.

\bibitem{llm_graphs}
\BIBentryALTinterwordspacing
B.~Jin, G.~Liu, C.~Han, M.~Jiang, H.~Ji, and J.~Han, ``Large language models on graphs: A comprehensive survey,'' 2023. [Online]. Available: \url{https://arxiv.org/abs/2312.02783}
\BIBentrySTDinterwordspacing

\bibitem{SanaSemEQ23}
M.~Sana and E.~C. Strinati, ``Semantic channel equalizer: Modelling language mismatch in multi-user semantic communications,'' in \emph{GLOBECOM 2023 - 2023 IEEE Global Communications Conference}, 2023, pp. 2221--2226.

\bibitem{fiorellino2024dynamic}
S.~Fiorellino, C.~Battiloro, E.~Calvanese~Strinati, and P.~Di~Lorenzo, ``Dynamic relative representations for goal-oriented semantic communications,'' \emph{arXiv preprint arXiv:2403.16986}, March 2024.

\bibitem{Petar2021semantic}
Q.~Lan \emph{et~al.}, ``What is semantic communication? a view on conveying meaning in the era of machine intelligence,'' \emph{Journal of Communications and Information Networks}, vol.~6, no.~4, pp. 336--371, 2021.

\bibitem{Calvanese2024goal}
E.~C. Strinati, P.~Di~Lorenzo \emph{et~al.}, ``Goal-oriented and semantic communication in {6G} ai-native networks: The {6G}-goals approach,'' \emph{arXiv preprint arXiv:2402.07573}, 2024.

\bibitem{b1}
L.~Hu, Y.~Li, H.~Zhang, L.~Yuan, F.~Zhou, and Q.~Wu, ``Robust semantic communication driven by knowledge graph,'' in \emph{2022 9th International Conference on Internet of Things: Systems, Management and Security (IOTSMS)}, 2022, pp. 1--5.

\bibitem{b3}
\BIBentryALTinterwordspacing
S.~Jiang \emph{et~al.}, ``Reliable semantic communication system enabled by knowledge graph,'' \emph{Entropy}, vol.~24, no.~6, p. 846, Jun. 2022. [Online]. Available: \url{http://dx.doi.org/10.3390/e24060846}
\BIBentrySTDinterwordspacing

\bibitem{b4}
F.~Zhou, Y.~Li, X.~Zhang, Q.~Wu, X.~Lei, and R.~Q. Hu, ``Cognitive semantic communication systems driven by knowledge graph,'' in \emph{ICC 2022 - IEEE International Conference on Communications}, 2022, pp. 4860--4865.

\bibitem{b5}
F.~Zhou \emph{et~al.}, ``Cognitive semantic communication systems driven by knowledge graph: Principle, implementation, and performance evaluation,'' \emph{IEEE Transactions on Communications}, vol.~72, no.~1, pp. 193--208, 2024.

\bibitem{morris2014writings}
C.~W. Morris, \emph{Writings on the general theory of signs}.\hskip 1em plus 0.5em minus 0.4em\relax Walter de Gruyter, 2014, vol.~16.

\bibitem{llm-kg}
\BIBentryALTinterwordspacing
M.~Trajanoska, R.~Stojanov, and D.~Trajanov, ``Enhancing knowledge graph construction using large language models,'' 2023. [Online]. Available: \url{https://arxiv.org/abs/2305.04676}
\BIBentrySTDinterwordspacing

\bibitem{gine}
\BIBentryALTinterwordspacing
W.~Hu \emph{et~al.}, ``Strategies for pre-training graph neural networks,'' 2019. [Online]. Available: \url{https://arxiv.org/abs/1905.12265}
\BIBentrySTDinterwordspacing

\bibitem{mine}
\BIBentryALTinterwordspacing
M.~I. Belghazi \emph{et~al.}, ``Mine: Mutual information neural estimation,'' 2018. [Online]. Available: \url{https://arxiv.org/abs/1801.04062}
\BIBentrySTDinterwordspacing

\bibitem{attention}
A.~Vaswani \emph{et~al.}, ``Attention is all you need,'' in \emph{Advances in Neural Information Processing Systems}, I.~Guyon, U.~V. Luxburg, S.~Bengio, H.~Wallach, R.~Fergus, S.~Vishwanathan, and R.~Garnett, Eds., vol.~30.\hskip 1em plus 0.5em minus 0.4em\relax Curran Associates, Inc., 2017.

\bibitem{webnlg}
\BIBentryALTinterwordspacing
C.~Gardent, A.~Shimorina, S.~Narayan, and L.~Perez-Beltrachini, ``The webnlg challenge: Generating text from rdf data,'' in \emph{Proceedings of the 10th International Conference on Natural Language Generation}.\hskip 1em plus 0.5em minus 0.4em\relax Association for Computational Linguistics, 2017. [Online]. Available: \url{http://dx.doi.org/10.18653/v1/W17-3518}
\BIBentrySTDinterwordspacing

\bibitem{bert-sentence}
\BIBentryALTinterwordspacing
N.~Reimers and I.~Gurevych, ``Sentence-bert: Sentence embeddings using siamese bert-networks,'' in \emph{Proceedings of the 2019 Conference on Empirical Methods in Natural Language Processing}.\hskip 1em plus 0.5em minus 0.4em\relax Association for Computational Linguistics, 11 2019. [Online]. Available: \url{https://arxiv.org/abs/1908.10084}
\BIBentrySTDinterwordspacing

\end{thebibliography}
\end{document}